\documentclass[10pt,letterpaper,twocolumn]{article}
\usepackage[margin=0.75in,columnsep=0.28in]{geometry}
\usepackage[T1]{fontenc}
\usepackage{newtxtext,newtxmath}
\usepackage[hyphens]{url}
\usepackage{graphicx}
\usepackage[round,authoryear]{natbib}
\usepackage{caption}
\usepackage{amsmath}
\usepackage{algorithm}
\usepackage{algorithmic}
\usepackage{newfloat}
\usepackage{listings}
\usepackage{booktabs}
\usepackage{xcolor}
\usepackage{microtype}

\DeclareCaptionStyle{ruled}{labelfont=normalfont,labelsep=colon,strut=off}
\floatstyle{ruled}
\newfloat{listing}{tb}{lst}{}
\floatname{listing}{Listing}

\title{\textbf{Lens: Bringing the Right Semantic Perspective into Focus for Training-Free Multimodal Representation Learning}}
\author{%
\begin{tabular}{c}
Xinran Liu,
Shouqian Shi\textsuperscript{\textdagger},
Yixian Chen,
Ruizhi Chen,\\
Xin-Wei Yao,
Sheng Zhong\\[0.35em]
\small \textsuperscript{\textdagger}Corresponding author: \texttt{sqlite@nju.edu.cn}
\end{tabular}%
}
\date{}

\begin{document}

\maketitle
\begin{abstract}
High-quality representations are essential for a wide range of
downstream tasks. Dedicated embedding models are explicitly optimized
for representation learning, yet their training data are often more
limited in scale and diversity than the massive corpora used to
pretrain modern large language models and multimodal large language
models. Large-scale pretraining and instruction following enable
autoregressive models to select relevant evidence, integrate
multimodal information, and infer semantics under different task
perspectives, creating a distinctive opportunity for training-free
representation learning.
However, our analysis reveals that existing semantic-elicitation
methods do not reliably orient the extracted states toward the semantic
perspective required by the downstream task. Consequently, the
resulting representations often remain dominated by salient input
content. We characterize this problem as \emph{semantic perspective
misalignment} and propose \textbf{Lens}, a training-free framework that
makes representation readout task-directed. \emph{Semantic Perspective Anchoring} associates the task-required
perspective with a task-specific readout phrase, specifying the
interpretive role of the positions later used for extraction. \emph{Contextualized Phrase Readout} places the same
phrase after the complete input and aggregates its token states,
combining full-context access with the anchored perspective. The
resulting representation reflects task-conditioned evidence integration
and inference rather than a generic summary of salient content. Without
parameter updates, architectural modification, or reranking, Lens
achieves an overall Precision@1 of 63.9 across all 36 MMEB datasets, outperforming the closest same-backbone training-free embedding
baseline by 10.2 points.
\end{abstract}

\section{Introduction}

Autoregressive models are increasingly being repurposed as
general-purpose representation models. Unlike dedicated embedding
encoders, large language models (LLMs) and multimodal large language
models (MLLMs) are pretrained to interpret content under diverse
instructions and contexts. Their hidden states can therefore reflect
not only what is explicitly present in an input, but also how the input
should be understood for a particular task. This ability makes frozen
autoregressive models a promising foundation for training-free
representation learning.

Recent studies have begun to convert decoder hidden states into
embeddings. Training-based approaches adapt pretrained models through
contrastive objectives, embedding supervision, or architectural
modification~\cite{behnamghader2024llm2vec,jiang2024vlm2vec,
lin2024mmembed,zhang2025gme}. Training-free approaches instead elicit
representations directly from frozen hidden states~\cite{jiang2024prompteol,lei2024metaeol,springer2024echo,
jiang2024e5v,zhu2026freeret}. In multimodal settings, E5-V introduces
semantic guidance for hidden-state extraction~\cite{jiang2024e5v},
while FreeRet further incorporates task instructions, noise suppression, and layer selection~\cite{zhu2026freeret}.
These methods demonstrate the potential of frozen decoders for direct
representation extraction without additional optimization.

Despite this progress, our analysis of representative failure cases
produced by a prior semantic-elicitation baseline reveals a consistent
pattern across tasks: the extracted representations capture plausible
input semantics, yet emphasize a perspective different from that
required for downstream comparison.

\begin{figure*}[t]
\centering
\includegraphics[width=0.80\textwidth]{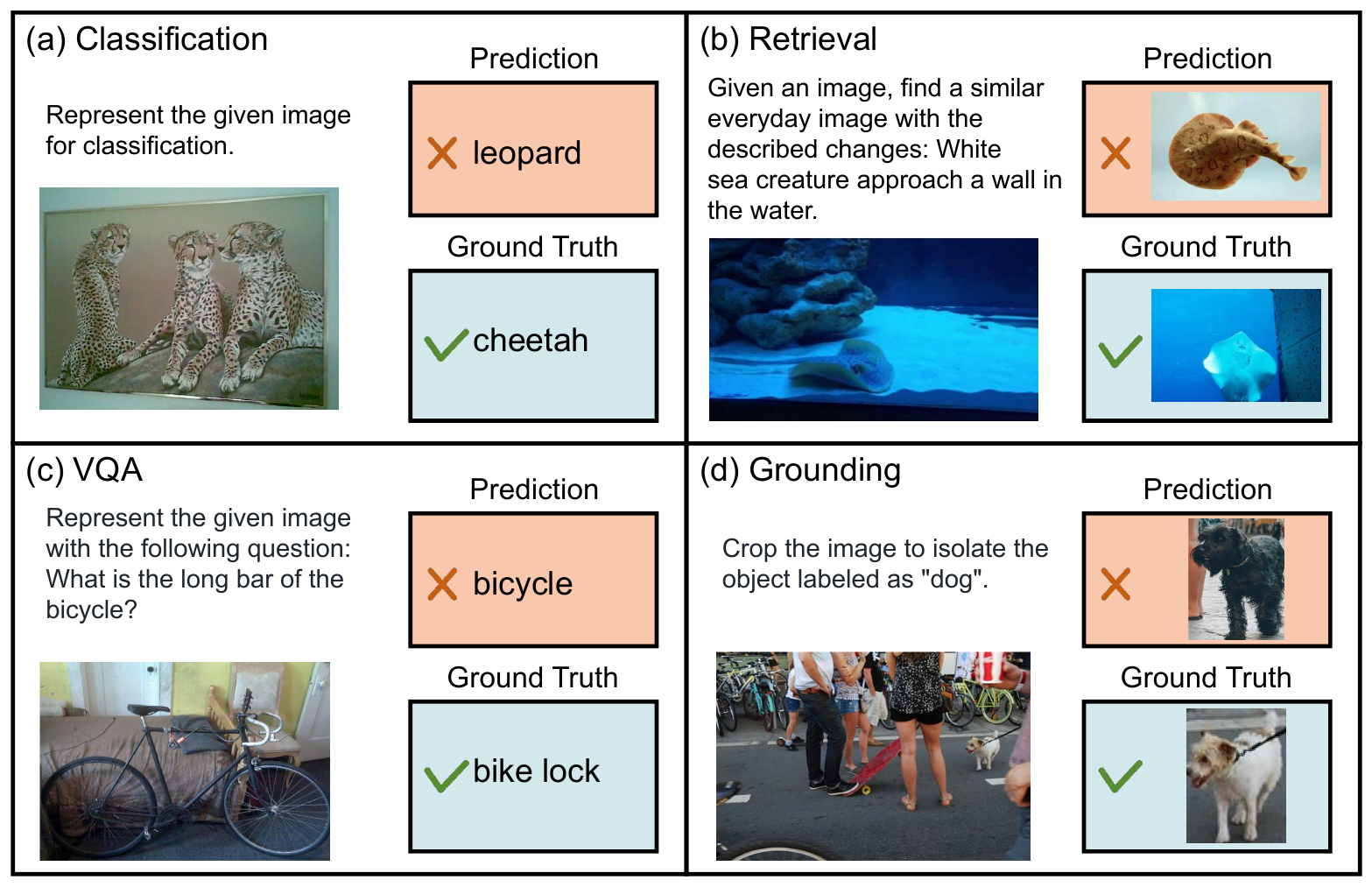}
\caption{
Representative failure cases produced by a prior
semantic-elicitation baseline across different downstream tasks.
}
\label{fig:observation}
\end{figure*}

As illustrated in Fig.~\ref{fig:observation}, classification requires
abstracting instance appearance into category identity; visual question
answering requires integrating the question with visual evidence to
derive the answer; and compositional retrieval requires resolving the
semantic effect of a specified modification. Yet the extracted
representations remain centered on immediately salient content. This
observed perspective mismatch reflects a deeper limitation: existing
readouts do not fully translate the decoder's task-conditioned
understanding into the resulting embedding. We define the mismatch
between the semantic perspective required by the task and that
expressed by the representation as \emph{semantic perspective
misalignment}.

To address this misalignment, we propose \textbf{Lens}, a training-free
framework that directs representation readout toward the required
semantic perspective. Lens associates a task-specific readout phrase
with this perspective and reuses the phrase after the complete task
specification and input. Its contextualized states combine an explicit
semantic role with the full preceding context, and their aggregation
yields a task-directed representation of the model's deeper
task-relevant understanding. Component ablations, controlled representation analysis, and
qualitative cases jointly support this mechanism, showing that Lens
redirects the readout toward the intended semantics and better captures
category abstraction, question-conditioned inference, and
compositional understanding. Lens
requires no parameter updates, architectural modification, or
reranking.

Our contributions are summarized as follows:
\begin{itemize}
    \item Through cross-task error analysis, we identify and
    characterize \emph{semantic perspective misalignment}, a recurring
    limitation in which the semantic perspective expressed by a
    representation differs from that required for downstream
    comparison.

    \item We introduce \textbf{Lens}, a general training-free framework
    that converts the task-conditioned understanding of frozen
    autoregressive models into task-directed and directly comparable
    representations, establishing a new perspective for exploiting
    their semantic and reasoning capabilities without additional
    optimization.

    \item We provide comprehensive empirical evidence on all 36 MMEB
    datasets spanning classification, visual question answering,
    retrieval, and grounding. Lens achieves an overall Precision@1 of 63.9 and outperforms the closest same-backbone training-free embedding baseline by 10.2 points. Component ablations, controlled representation analysis, and qualitative diagnostics further support the proposed mechanism and
demonstrate its effectiveness.
\end{itemize}
\section{Related Work}

\subsection{Multimodal Representation Learning}

Multimodal representation learning maps heterogeneous inputs into an
embedding space that supports efficient cross-modal comparison.
Dual-encoder models such as CLIP~\cite{radford2021learning},
ALIGN~\cite{jia2021scaling}, and SigLIP~\cite{zhai2023sigmoid} learn
transferable vision--language representations from large-scale paired
data. More recent methods repurpose multimodal large language models
(MLLMs) as general-purpose embedding models. VLM2Vec introduces the
Massive Multimodal Embedding Benchmark and contrastively adapts MLLMs
across diverse embedding tasks~\cite{jiang2024vlm2vec}. MM-Embed
improves universal multimodal retrieval through modality-aware
hard-negative mining~\cite{lin2024mmembed}, while GME explores
large-scale training for general multimodal embeddings~\cite{zhang2025gme}. These approaches obtain strong representations
through additional optimization. Lens instead extracts task-directed
representations directly from a frozen multimodal decoder.

\subsection{Training-Free Representations from Autoregressive Models}

Autoregressive models have also been adapted for representation
learning without parameter updates. In the text domain, PromptEOL uses
natural-language prompts to elicit sentence representations~\cite{jiang2024prompteol}, while MetaEOL combines multiple elicited
views to capture complementary semantics~\cite{lei2024metaeol}. Echo
embeddings repeat the input so that later token states receive fuller
context under causal attention~\cite{springer2024echo}. These methods
demonstrate that prompt design and readout position substantially affect
the representational quality of frozen decoders.

Training-free multimodal representation learning extends this idea to
MLLMs. E5-V derives embeddings from semantically elicited hidden states~\cite{jiang2024e5v}. FreeRet further improves task alignment, semantic
grounding, noise suppression, and hidden-state selection, and employs
the frozen model for retrieval and reranking~\cite{zhu2026freeret}.
These methods primarily improve how meaningful semantic states are
elicited and selected. Lens addresses a complementary limitation:
whether the selected states express the semantic perspective required
for downstream comparison.

\begin{figure*}[t]
    \centering
    \includegraphics[width=0.98\textwidth]{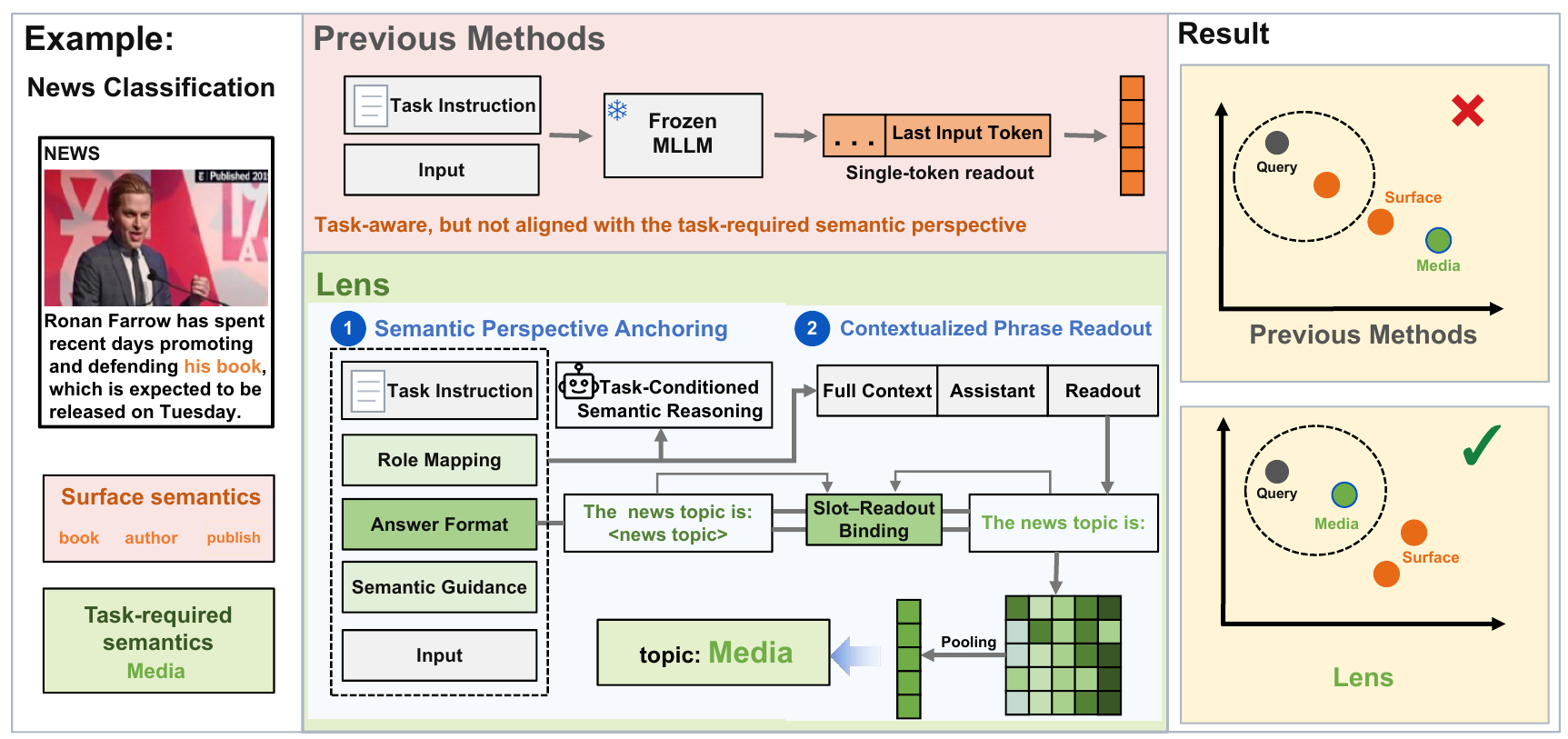}
    \caption{
    Overview of Lens. It anchors the task-required perspective to a
    readout phrase and aggregates the phrase's full-context states as
    the representation.
    }
    \label{fig:method}
\end{figure*}

\section{Method}
\label{sec:method}

\subsection{Overview}
\label{sec:method_overview}

Given a downstream task $\tau$, Lens independently encodes a query
$q$ and a candidate $c$ with a frozen multimodal autoregressive model,
producing representations whose similarity reflects task-defined
compatibility.

\medskip
\noindent
\begin{minipage}{\columnwidth}
\hrule
\smallskip

\textit{User-side prompt}
\smallskip

\{Task Instruction\}\par
\{Role Mapping\}\par
\textbf{Answer format:}\par
\{Readout Phrase\}
\textless\{Task-Relevant Semantics\}\textgreater\par
\{Semantic Guidance\}\par
\{Input Content\}

\smallskip
\hrule
\smallskip

\textit{Assistant-side prefill}
\smallskip

\{Readout Phrase\}

\smallskip
\hrule
\end{minipage}
\medskip

Braced expressions are instantiated separately for each task and for
the query and candidate. The italicized headers distinguish the user
and assistant sides and are not included in the model input. Complete task-specific query and candidate prompts are provided in the supplementary material.

As illustrated in Fig.~\ref{fig:method}, Lens consists of two coupled
components. \emph{Semantic Perspective Anchoring} establishes the
semantics with respect to which the query and candidate should be
compared. \emph{Contextualized Phrase Readout} extracts the
input-specific realization of those semantics from the model's internal
states. The former determines what the representation should express,
while the latter determines where that representation is read.

The task-required semantic perspective is the task-specific semantic
content on which query--candidate compatibility is judged. The two
sides may derive or express this content differently, but their
representations must organize it in a directly comparable form. The
task instruction defines the compatibility relation, while role mapping
specifies how each side contributes the required semantics. The answer
format associates these semantics with a shared readout phrase, and
semantic guidance specifies how the current input should be interpreted
under the resulting perspective.

For news classification, the query-side prompt derives the topic
conveyed by an article, whereas the candidate-side prompt represents
the meaning of a topic category. Both sides share the answer format
``The news topic is:
\textless news topic\textgreater'' and the readout phrase
``The news topic is:'', while their role mappings and semantic guidance
are instantiated separately.

The two occurrences of the readout phrase serve complementary roles.
Its occurrence in the answer format establishes its association with
the task-required semantic perspective. Its assistant-side occurrence
follows the complete task and input context and supplies the hidden
states used to construct the representation. Lens thus couples semantic
specification and contextual readout through the same linguistic phrase.

\subsection{Semantic Perspective Anchoring}
\label{sec:semantic_anchoring}

Representation similarity is meaningful only when it reflects the
semantics on which task compatibility depends. A representation may
capture salient and plausible properties of an input yet remain
unsuitable for comparison if those properties are not the semantics
under which the query and candidate should be matched. Semantic
Perspective Anchoring addresses this mismatch by establishing the
task-required perspective that representation similarity should
reflect.

For a given task, this perspective must apply to both input sides while
respecting their distinct informational roles. A query may require
inference from multimodal context, whereas a candidate may directly
express the relevant semantics. Role mapping specifies this distinction
while organizing both sides around the same perspective.

Lens associates the task-required perspective with a shared readout
phrase through the answer format:
\[
\text{\{Readout Phrase\}
\textless\{Task-Relevant Semantics\}\textgreater}.
\]
The complete declaration appears before the input and places the phrase
and the task-relevant semantics in an explicit linguistic relation. The
first occurrence of the phrase is not used as the representation.
Instead, the complete answer format becomes part of the preceding
context available to the later assistant-side occurrence.

The anchored perspective must correspond to the semantics that govern
task compatibility rather than to an unrestricted description of
either input. It must remain consistent across the query and candidate,
even when their roles differ, and the same readout phrase must be used
in the answer format and assistant-side prefill. These conditions place
both representations under a common task-defined interpretation.

Semantic guidance specifies how the task-required perspective should be
derived from each input side. It directs the model to organize the
relevant evidence, perform the required abstraction or inference, and
preserve the semantic distinctions needed for task compatibility.
Unlike the answer format, which defines the shared perspective,
semantic guidance may differ between the query and candidate according
to their informational roles.

By establishing an explicit association between the readout phrase and
the task-required perspective, Semantic Perspective Anchoring targets
semantic perspective misalignment at its source. It also yields a
testable prediction: modifying or removing this association should
systematically redirect the resulting representation, as examined in
Table~\ref{tab:perspective_analysis}.

\subsection{Contextualized Phrase Readout}
\label{sec:phrase_readout}

Semantic Perspective Anchoring establishes the interpretation
associated with the readout phrase. Contextualized Phrase Readout places
the same phrase after the complete prompt and extracts its token states.
Because these tokens occur after the task instruction, role mapping,
answer format, semantic guidance, and input, their hidden states are
conditioned jointly on the anchored perspective and the current input
context.

The resulting states encode the current input at the readout phrase
whose semantic role has been established by the preceding context.
Lens extracts the states of the phrase itself without generating an
answer or reading any subsequently generated content.

Suppose that the readout phrase contains $m$ tokens at positions
$\{p_1,\ldots,p_m\}$. For each side $s\in\{q,c\}$, Lens extracts their
final-layer states after the self-attention residual addition and before
the post-attention layer normalization and MLP. The representation is
computed as
\begin{equation}
\widetilde{\mathbf{e}}_{\tau}^{s}
=
\frac{1}{m}
\sum_{j=1}^{m}
\mathbf{h}_{L,p_j}^{s},
\qquad
\mathbf{e}_{\tau}^{s}
=
\frac{
\widetilde{\mathbf{e}}_{\tau}^{s}
}{
\left\|
\widetilde{\mathbf{e}}_{\tau}^{s}
\right\|_2
},
\label{eq:phrase_readout}
\end{equation}
where $\mathbf{h}_{L,p_j}^{s}$ denotes the hidden state of the $j$-th
token in the assistant-side readout phrase.

Lens aggregates all phrase-token states because the semantic role is
assigned to the readout phrase as a whole rather than to any individual
token. Averaging its contextualized token states avoids privileging an
arbitrary lexical position and retains the phrase-level realization of
the anchored perspective.

For task $\tau$, query--candidate compatibility is measured by cosine
similarity:
\begin{equation}
\operatorname{sim}_{\tau}(q,c)
=
\left(\mathbf{e}_{\tau}^{q}\right)^{\top}
\mathbf{e}_{\tau}^{c}.
\label{eq:similarity}
\end{equation}

Query and candidate inputs are encoded independently, allowing
candidate representations to be precomputed and reused. Lens requires
one forward pass per input, without parameter updates, architectural
modification, answer generation, or reranking.

\section{Experiments}
\label{sec:experiments}

\subsection{Experimental Setup}

\paragraph{Benchmark and Evaluation Protocol.}

We evaluate Lens on the Massive Multimodal Embedding Benchmark
(MMEB)~\cite{jiang2024vlm2vec}, which contains 36 datasets spanning
10 classification, 10 visual question answering (VQA), 12 retrieval,
and 4 grounding datasets. We select MMEB because its unified
candidate-ranking protocol covers substantially different forms of
multimodal compatibility, allowing us to evaluate whether Lens
generalizes across task semantics without task-specific training.
Following the standard MMEB protocol, all tasks are formulated as
candidate-ranking problems, and we report Precision@1 (\%).
Task-family scores are averaged over their constituent datasets, while
the overall average is computed across all 36 datasets.
\paragraph{Baselines.}

We compare Lens with three groups of multimodal representation methods.
\emph{Conventional embedding models} include
CLIP~\cite{radford2021learning},
BLIP-2~\cite{li2023blip2},
SigLIP~\cite{zhai2023sigmoid},
OpenCLIP~\cite{cherti2023reproducible}, and
MagicLens~\cite{zhang2024magiclens}.
\emph{Training-based MLLM embedding methods} include
VLM2Vec~\cite{jiang2024vlm2vec},
UniME~\cite{gu2025unime},
MMRet~\cite{zhou2024mmret},
MM-Embed~\cite{lin2024mmembed},
LamRA-Ret~\cite{liu2025lamra}, and
GME~\cite{zhang2025gme}.
\emph{Training-free MLLM embedding methods} include
E5-V~\cite{jiang2024e5v} and
FreeRet~\cite{zhu2026freeret}. We report the embedding-only variant of
FreeRet, denoted FreeRet-embed, without reranking. Its
Qwen2.5-VL-7B result provides the closest same-backbone comparison.

Results for the conventional embedding models and VLM2Vec are taken
from VLM2Vec~\cite{jiang2024vlm2vec}. The remaining baseline results
are taken from FreeRet~\cite{zhu2026freeret}.

\paragraph{Implementation Details.}

We use Qwen2.5-VL-7B-Instruct~\cite{bai2025qwen25vl} as the primary
frozen backbone and retain the original FreeRet task instructions
across all 36 MMEB datasets for controlled comparison. We further
evaluate Qwen2.5-VL-3B-Instruct~\cite{bai2025qwen25vl} and
Qwen3.5-4B/9B~\cite{qwen35} without backbone-specific adaptation.
Complete task-specific query and candidate prompts are provided in the
supplementary material.

For each input, we extract the final-layer states of the assistant-side
readout phrase after the self-attention residual addition and before the
post-attention layer normalization and MLP. Phrase-token states are
mean-pooled and $\ell_2$-normalized. Queries and candidates are encoded
independently in a single forward pass, allowing candidate
representations to be precomputed and reused.

Lens requires no parameter updates, additional training data,
architectural modification, answer generation, or reranking.
\subsection{Main Results}

\begin{table*}[t]
\centering
{
\small
\setlength{\tabcolsep}{4.0pt}
\begin{tabular}{@{}llccccc@{}}
\toprule
\textbf{Method}
& \textbf{Backbone}
& \textbf{Classification}
& \textbf{VQA}
& \textbf{Retrieval}
& \textbf{Grounding}
& \textbf{Average} \\
\midrule

\multicolumn{7}{@{}l}{
\textit{Conventional Multimodal Embedding Models}} \\

CLIP~\cite{radford2021learning}
& --
& 42.8
& 9.1
& 53.0
& 51.8
& 37.8 \\

BLIP-2~\cite{li2023blip2}
& --
& 27.0
& 4.2
& 33.9
& 47.0
& 25.2 \\

SigLIP~\cite{zhai2023sigmoid}
& --
& 40.3
& 8.4
& 31.6
& 59.5
& 34.8 \\

OpenCLIP~\cite{cherti2023reproducible}
& --
& 47.8
& 10.9
& 52.3
& 53.3
& 39.7 \\

MagicLens~\cite{zhang2024magiclens}
& --
& 38.8
& 8.3
& 35.4
& 26.0
& 27.8 \\

\midrule
\multicolumn{7}{@{}l}{
\textit{Training-Based MLLM Embedding Methods}} \\

VLM2Vec~\cite{jiang2024vlm2vec}
& LLaVA-1.6-7B
& 61.2
& 49.9
& 67.4
& \underline{86.1}
& 62.9 \\

UniME~\cite{gu2025unime}
& LLaVA-1.6-7B
& 43.0
& 17.7
& 42.5
& 63.2
& 41.6 \\

MMRet~\cite{zhou2024mmret}
& LLaVA-1.6-7B
& 47.2
& 18.4
& 56.5
& 62.2
& 44.0 \\

MM-Embed~\cite{lin2024mmembed}
& LLaVA-Next-7B
& 48.1
& 32.3
& 63.8
& 57.8
& 50.0 \\

LamRA-Ret~\cite{liu2025lamra}
& Qwen2.5-VL-7B
& 51.7
& 34.1
& 66.9
& 56.7
& 52.4 \\

GME~\cite{zhang2025gme}
& Qwen2-VL-7B
& 57.7
& 34.7
& \underline{71.2}
& 59.3
& 56.0 \\

\midrule
\multicolumn{7}{@{}l}{
\textit{Training-Free MLLM Embedding Methods}} \\

E5-V$^\dagger$~\cite{jiang2024e5v}
& Qwen2.5-VL-7B
& 41.2
& 37.2
& 37.9
& 48.4
& 39.8 \\

FreeRet-embed~\cite{zhu2026freeret}
& LLaVA-OV-7B
& 53.0
& 47.4
& 45.7
& 53.6
& 49.1 \\

FreeRet-embed~\cite{zhu2026freeret}
& Qwen2-VL-7B
& 59.0
& 50.2
& 52.3
& 60.1
& 54.5 \\

FreeRet-embed~\cite{zhu2026freeret}
& Qwen2.5-VL-7B
& 59.7
& 52.8
& 49.2
& 54.7
& 53.7 \\

Lens (Ours)
& Qwen2.5-VL-3B
& 56.9
& 51.6
& 51.6
& 58.8
& 53.9 \\

\textbf{Lens (Ours)}
& Qwen2.5-VL-7B
& \textbf{\underline{70.3}}
& \textbf{\underline{63.4}}
& \textbf{59.6}
& \textbf{61.7}
& \textbf{\underline{63.9}} \\

\bottomrule
\end{tabular}
}
\caption{
Main results on MMEB in Precision@1 (\%). The classification, VQA,
retrieval, and grounding columns report averages over 10, 10, 12,
and 4 datasets, respectively. Baseline results, including the reported
averages, follow the corresponding sources: conventional embedding
models and VLM2Vec are taken from VLM2Vec~\cite{jiang2024vlm2vec},
while the remaining baselines are taken from
FreeRet~\cite{zhu2026freeret}. Boldface denotes the best result among
training-free methods, while underlining denotes the best result among
all methods shown. A result satisfying both criteria is both boldfaced
and underlined. $^\dagger$ E5-V is reproduced by FreeRet with
Qwen2.5-VL-7B.
}
\label{tab:main_result}
\end{table*}

Table~\ref{tab:main_result} reports the average performance across the
four MMEB task families. Lens with Qwen2.5-VL-7B achieves the best
training-free result in every task family and reaches an overall
Precision@1 of 63.9. Compared with FreeRet-embed using the same
backbone, Lens improves the overall average by 10.2 points, with gains
of 10.6, 10.6, 10.4, and 7.0 points on classification, VQA, retrieval,
and grounding, respectively.

Lens obtains particularly strong results on classification and VQA,
where successful matching depends on identifying the specific label or
answer semantics required by the query, directly benefiting from the
task-oriented representations constructed by Lens. The substantial gain
on retrieval further shows that aligning heterogeneous inputs under a
shared semantic perspective improves global semantic matching. The gain
on grounding is comparatively smaller because region candidates already
provide strong localized visual cues, leaving less room for semantic
reorganization; nevertheless, Lens still achieves the best
training-free result by directing these cues toward the relation
specified by the query.

\paragraph{Different Backbones and Model Sizes.}

With Qwen2.5-VL-3B, Lens achieves an overall Precision@1 of 53.9,
comparable to the 53.7 obtained by FreeRet-embed with the larger
Qwen2.5-VL-7B backbone. Lens further obtains overall scores of 53.2
and 55.2 with Qwen3.5-4B and Qwen3.5-9B, respectively. Their
classification, VQA, retrieval, and grounding scores are
57.1/54.8/45.8/61.8 and 60.0/55.9/47.4/65.1, respectively. These
results show that the same Lens formulation can be applied across
different multimodal model families and scales without
backbone-specific adaptation.

Qwen2.5-VL-7B nevertheless achieves the strongest overall performance,
indicating that the effectiveness of training-free readout depends not
only on model size, but also on how reliably the backbone exposes
task-conditioned multimodal semantics at the selected internal
position.

\subsection{Ablation Studies}
\begin{table*}[t]
\centering
{
\small
\setlength{\tabcolsep}{5.0pt}
\begin{tabular}{@{}lccccc@{}}
\toprule
\textbf{Variant}
& \textbf{Classification}
& \textbf{VQA}
& \textbf{Retrieval}
& \textbf{Grounding}
& \textbf{Average} \\
\midrule

Full Lens
& \textbf{70.3}
& \textbf{63.4}
& \textbf{59.6}
& 61.7
& \textbf{63.9} \\

w/o Answer-Format Anchoring
& 61.6
& 58.4
& 55.0
& 64.4
& 58.8 \\

Last-Token Readout
& 61.4
& 54.2
& 48.6
& \textbf{65.8}
& 55.6 \\

\bottomrule
\end{tabular}
}
\caption{
Ablation results on MMEB in Precision@1 (\%). The first variant removes
the answer-format declaration that associates the task-required
perspective with the readout phrase. The second retains the complete
prompt and assistant-side phrase but replaces phrase-level aggregation
with the last phrase-token state. All other settings remain unchanged.
The best result in each column is boldfaced.
}
\label{tab:ablation}
\end{table*}

Table~\ref{tab:ablation} evaluates the two defining operations of Lens
across all 36 MMEB datasets. Removing answer-format anchoring reduces
the overall Precision@1 by 5.1 points, while replacing phrase-level
aggregation with the last phrase-token state causes an 8.3-point
decrease, demonstrating that both operations are important to the broad
effectiveness of Lens. The two ablated variants obtain higher grounding
scores, where the candidates are already localized image regions and
successful matching depends more directly on preserving fine-grained
local visual correspondence. In this setting, weaker task-level
semantic reorganization may retain more of the backbone's native local
visual evidence. Nevertheless, the complete design provides
substantially stronger classification, VQA, and retrieval results and
achieves the highest overall performance.

\paragraph{Semantic Perspective Anchoring.}

Removing the answer-format anchor lowers the overall Precision@1 from
63.9 to 58.8. This variant retains the task instruction, role mapping,
semantic guidance, and assistant-side readout phrase, but no longer
explicitly associates the phrase with the semantic perspective required
by the task. This causes decreases of 8.7, 5.0, and 4.6 points on
classification, VQA, and retrieval, respectively. These results show
that merely providing the relevant task information is insufficient:
explicitly anchoring the readout phrase to the task-required perspective
is important for organizing heterogeneous queries and candidates under
a consistent basis of comparison.

\paragraph{Contextualized Phrase Readout.}

Using only the last phrase-token state reduces the overall Precision@1
from 63.9 to 55.6 while retaining the answer-format anchor, complete
prompt, and assistant-side phrase. It produces decreases of 8.9, 9.2,
and 11.0 points on classification, VQA, and retrieval, respectively.
This substantial degradation indicates that the task-relevant evidence
organized at the readout phrase is distributed across its
contextualized token states rather than being fully represented by a
single lexical position. Aggregating the complete phrase therefore
preserves a more comprehensive task-directed representation for
subsequent similarity comparison.

\begin{table}[t]
\centering
{
\small
\begin{tabular}{@{}p{0.72\columnwidth}r@{}}
\toprule
\textbf{Comparison}
& \textbf{Similarity} \\
\midrule

Task-oriented $\leftrightarrow$ correct topic
& 0.981 \\

Content-oriented $\leftrightarrow$ correct topic
& 0.965 \\

Unanchored $\leftrightarrow$ correct topic
& 0.960 \\

\midrule

Unanchored $\leftrightarrow$ content-oriented
& 0.998 \\

Unanchored $\leftrightarrow$ task-oriented
& 0.983 \\

\bottomrule
\end{tabular}
}
\caption{
Average cosine similarities on the N24News test split. The first block
compares each query representation with its correct topic
representation; the second compares the unanchored representation with
its content-oriented and task-oriented counterparts.
}
\label{tab:perspective_analysis}
\end{table}

\begin{figure*}[t]
\centering
\includegraphics[width=\textwidth]{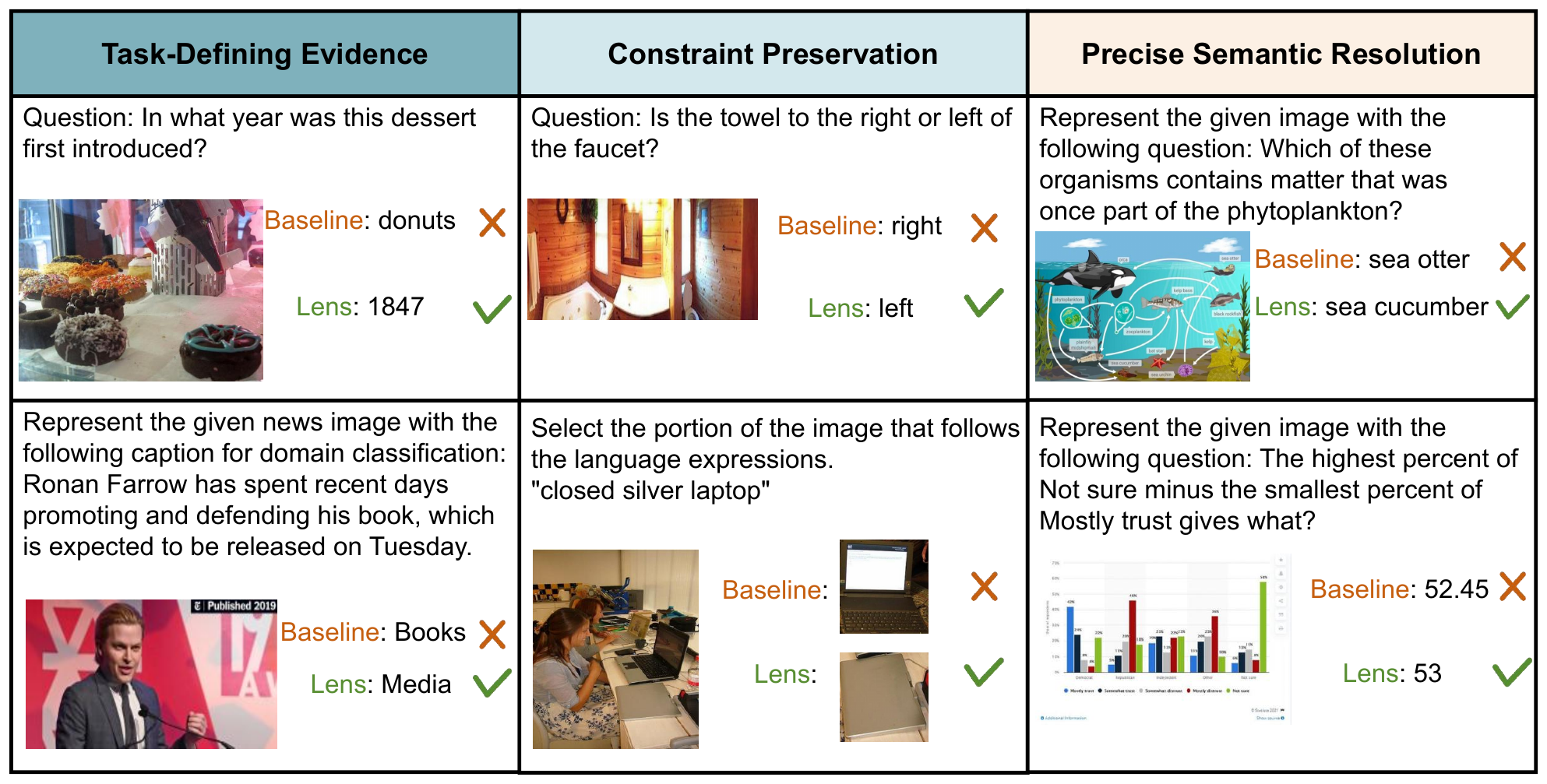}
\caption{
Representative cases corrected by Lens, illustrating
question-conditioned inference, compositional-constraint preservation,
and knowledge-supported inference.
}
\label{fig:discussion}
\end{figure*}

\subsection{Representation-Level Effect of Semantic Perspective
Anchoring}
\label{sec:perspective_analysis}

We further examine whether the answer format can control the semantic
perspective expressed at the readout phrase. We conduct the analysis on
the full N24News test split~\cite{wang2022n24news} using the frozen
Qwen2.5-VL-7B backbone. Candidate topic representations are encoded
once with the task-oriented target prompt and remain fixed across all
query conditions. We construct three query conditions while keeping the input, task instruction, semantic guidance, assistant-side readout phrase, and
extraction procedure unchanged. The \emph{task-oriented} condition uses
``The news topic is: \textless news topic\textgreater'' as the answer
format, the \emph{content-oriented} condition replaces it with
``The input content is: \textless input content\textgreater'', and the
\emph{unanchored} condition removes the answer format. Thus, the
experiment tests whether changing only the semantic perspective
specified by the answer format redirects the representation extracted
from the same readout phrase.

As shown in Table~\ref{tab:perspective_analysis}, the task-oriented
representation has the highest similarity to the correct topic
representation, whereas the content-oriented and unanchored variants
are less aligned with the task target. Moreover, the unanchored
representation is closer to the content-oriented representation than
to the task-oriented one. These results show that changing the answer
format redirects the semantic perspective expressed at the fixed
readout phrase: task-oriented anchoring moves the representation toward
the news-topic semantics required for comparison, whereas changing or
removing the anchor shifts it away from this task-oriented state. This
directly supports the central mechanism of Semantic Perspective
Anchoring.

\subsection{Qualitative Analysis of Corrected Cases}
\label{sec:qualitative_analysis}

Figure~\ref{fig:discussion} presents representative cases that are
incorrectly predicted by FreeRet-embed but correctly resolved by Lens
under the same Qwen2.5-VL-7B backbone. Since the input, candidate set,
and frozen model remain unchanged, the differences reflect how the two
methods organize the backbone's internal semantics for comparison.
Across the examples, FreeRet-embed tends to emphasize globally salient
content, whereas Lens focuses on the evidence required by the query.

In the donut example, the task requires a historical relation rather
than object identity, and Lens shifts the representation toward the
queried year \emph{1847}. The \emph{left} and
\emph{closed silver laptop} cases similarly require preserving spatial
relations and compositional constraints beyond broad object similarity.
The ScienceQA cases further suggest that Lens can organize not only
explicit visual evidence, but also task-relevant knowledge and
reasoning encoded in the frozen backbone when the answer is not directly
stated in the input.

These examples illustrate semantic perspective misalignment: the same
input can support multiple plausible interpretations, but only one
provides the appropriate basis for candidate comparison. Semantic
Perspective Anchoring specifies this task-required interpretation,
while Contextualized Phrase Readout preserves the corresponding
evidence for direct comparison. Lens therefore goes beyond reorganizing
surface input content, providing a training-free way to expose and reuse
the broader knowledge and reasoning capabilities already acquired by
multimodal autoregressive models.

\section{Conclusion}

We introduced \textbf{Lens}, a training-free framework for deriving
multimodal representations from frozen autoregressive models. Through
cross-task error analysis, we identified \emph{semantic perspective
misalignment}: existing readouts can capture plausible input semantics
while failing to express the perspective required for downstream
comparison. Lens addresses this problem by explicitly associating the
task-required perspective with a readout phrase and extracting its
contextualized states after the complete input.
Without parameter updates, architectural modification, or reranking,
Lens achieves an overall Precision@1 of 63.9 across all 36 MMEB
datasets, outperforming the closest same-backbone training-free embedding
baseline by 10.2 points. Component ablations demonstrate the contributions of
Semantic Perspective Anchoring and Contextualized Phrase Readout, while
controlled representation analysis shows that anchoring redirects the
readout toward the intended semantic perspective. Qualitative cases
further show improvements in question-conditioned inference,
compositional understanding, and knowledge-supported reasoning. These
findings support a broader view of training-free representation
learning: embeddings need not be limited to compressed descriptions of
observed content, but can instead reflect the task-conditioned
semantics that a frozen multimodal decoder organizes, integrates, and
infers.

\bibliographystyle{plainnat}
\bibliography{references}

\end{document}